\documentclass[conference]{IEEEtran}
\IEEEoverridecommandlockouts
\usepackage{cite}
\usepackage{amsmath,amssymb,amsfonts}
\usepackage{algorithmic}
\usepackage{graphicx}
\usepackage{textcomp}
\usepackage{subcaption}
\usepackage{dblfloatfix}
\usepackage{xcolor}
\def\BibTeX{{\rm B\kern-.05em{\sc i\kern-.025em b}\kern-.08em
    T\kern-.1667em\lower.7ex\hbox{E}\kern-.125emX}}
\begin{document}

\title{Calibrating Urban Traffic Simulation from Sparse Road Observations via Genetic Optimization\\

}

\author{
\IEEEauthorblockN{Hunter Sawyer}
\IEEEauthorblockA{\textit{Department of Computer Science} \\
\textit{Tennessee Technological University}\\
Cookeville, TN \\
htsawyer42@tntech.edu}
\and
\IEEEauthorblockN{Jesse Roberts}
\IEEEauthorblockA{\textit{Department of Computer Science} \\
\textit{Tennessee Technological University}\\
Cookeville, TN \\
JTRoberts@tntech.edu}
\and
\IEEEauthorblockN{Simon Matei}
\IEEEauthorblockA{\textit{Department of Computer Science} \\
\textit{Tennessee Technological University}\\
Cookeville, TN \\
ssmatei42@tntech.edu}
}

\maketitle

\begin{abstract}
Urban traffic simulation is a critical tool for infrastructure planning, including the placement of electric vehicle charging stations. However, realistic traffic simulation across many cities is hindered by two fundamental data limitations: detailed real-world traffic measurements are available for only a small fraction of road segments in most cities, and employment distribution data critical for modeling commuter traffic is rarely available at the resolution needed for simulation. This paper presents a genetic algorithm-based framework that directly addresses both limitations, calibrating urban traffic simulations from sparse road observations without requiring detailed job location data. Using the SUMO traffic simulation platform for Greensboro, North Carolina, our approach optimizes job distributions and gate-traffic parameters to align simulated traffic with a small sample of roads with known traffic-flow rates. We demonstrate that this approach produces simulated traffic that correlates well with real-world measurements, generalizes to road segments withheld from training, and produces job distributions that show promising qualitative agreement with census employment data despite never directly training on that employment data. This work demonstrates that realistic urban traffic simulation can be achieved from minimal real-world observations, offering a scalable and data-light approach to simulation calibration that reduces the barrier to deploying traffic models across diverse cities.
\end{abstract}


\section{Introduction}
\label{sec:introduction}

Realistic urban traffic simulation is a critical enabler of infrastructure planning, from optimizing road networks to supporting the deployment of electric vehicle charging stations. However, deploying accurate traffic simulations across many cities faces two fundamental data challenges. First, comprehensive real-world traffic measurements are prohibitively expensive to collect, and in practice only a small fraction of road segments in any city have known traffic-flow rates. Second, the employment distribution data needed to model commuter traffic realistically is rarely available at the resolution required for simulation. Together, these limitations make it difficult to build traffic simulations that are both realistic and scalable across diverse urban environments.

Existing approaches to synthetic traffic generation, including rule-based simulations and evolutionary algorithms, have made significant progress but leave a critical gap. Much of this work focuses on calibrating microscopic driver behavior parameters such as reaction times, gap distances, and acceleration profiles, or assumes access to detailed demographic and employment data that is not broadly available. Comparatively little attention has been paid to calibrating the underlying spatial distribution of human activity that fundamentally drives traffic generation, and doing so from the kind of sparse observational data that is realistically available for most cities.

This paper addresses that gap directly. We present a genetic algorithm-based framework that calibrates urban traffic simulation from sparse road observations without requiring detailed employment distribution data, and show that optimizing against this sparse signal generalizes beyond the roads used for training, improving alignment on randomly held-out road segments and in many cases on geographically excluded regions of the city. As a secondary finding, recovered job distributions show promising qualitative agreement with census employment data despite never directly training on that data, suggesting that unavailable urban structure data may be partially recoverable from latent components within the traffic signal.

We evaluate our approach using the SUMO traffic simulation platform for Greensboro, North Carolina, addressing the following research questions:

\begin{enumerate}
    \item Does the genetic approach improve the alignment of simulated traffic with real-world traffic observations?
    \item Do randomly held-out road segments improve in alignment alongside the roads used for training?
    \item Do geographically excluded road segments improve in alignment alongside the roads used for training?
    \item Do the job distributions recovered by our framework show agreement with real-world employment data?
\end{enumerate}

Our results demonstrate that realistic urban traffic simulation can be achieved from minimal real-world observations, offering a scalable and data-light approach to simulation calibration that reduces the barrier to deploying traffic models across diverse cities.


\section{Background \& Related Work}

To contextualize the proposed methodology, this section first details the foundational simulation infrastructure and activity-based demand models utilized in our experiment, followed by a review of existing traffic calibration literature to highlight the critical gap in spatial demand optimization that our framework resolves.

\subsection{Background}
The genetic algorithm (GA) is a stochastic, metaheuristic optimization technique within the broader class of evolutionary algorithms in computer science. Inspired by the principles of biological evolution and natural selection, GAs are widely utilized to efficiently navigate complex search spaces and identify high-quality solutions to optimization problems. The algorithm evaluates and manipulates candidate parameters through biologically inspired operators such as selection, crossover, and mutation to iteratively evolve toward an optimal solution. Each potential solution being mutated is also known as a chromosome. In the context of urban planning and traffic generation, this iterative search algorithm is well-suited to solving optimization problems involving the calibration of crucial parameters that govern driver behavior and vehicular movement. However, we propose a novel application for utilizing GAs. Our GA modifies the spatial distribution of job placements while initializing residential locations from known housing data. To accurately evaluate how these dynamic spatial configurations impact overall traffic flow, our optimization model is paired with a high-fidelity microscopic traffic simulation known as the Simulation of Urban MObility (SUMO).

SUMO is an open-source, highly portable, microscopic traffic simulation package that serves as the basis for our simulation. The package is designed to handle large-scale road networks and provides the computational infrastructure to model individual vehicle movement and routing behaviors. Rather than constructing a theoretical road network from scratch to utilize within this simulator, our methodology employs the GreenEVT framework \cite{nilsson2024greenevt} as its foundational architecture. While originally developed as a co-simulation between electric distribution systems and traffic dynamics, GreenEVT provides a thoroughly validated, high-resolution urban road network. Utilizing this established framework supplies our simulation with a realistic geographic and infrastructural backdrop. However, GreenEVT was designed as an evacuation simulation to evaluate the effects electric vehicles have on the power grid, which is why we utilize another SUMO tool for realistic traffic generation, known as ActivityGen, to construct baseline daily traffic demand. ActivityGen generates a realistic traffic demand based on a simple activity-based traffic model supporting activities such as work, school, and free time. The locations of the work activities serve as foundational parameters that our GA iteratively optimizes to improve overall traffic flow.

\subsection{Related Work}
Traditionally, calibration of the parameters relating to traffic generation has been done through rigid, manual techniques that are labor-intensive and time-consuming. In large-scale simulations such as the Luxembourg SUMO Traffic (LuST) Scenario, these methods provide robust results but rely heavily on extensive real-world datasets and the manual analysis of road conditions and traffic demand \cite{codeca2015luxembourg}. To mitigate this manual burden, subsequent research has attempted to introduce algorithmic assistance. For instance, the Tokyo SUMO Traffic (ToST) scenario utilized a binary search algorithm to automatically adjust pass-through vehicle volumes, compensating for the rigidities of demographic demand data \cite{yamazaki2023tost}. This approach works to minimize an error function by adjusting the ActivityGen parameter for random trips that are not related to work or school. To fully overcome the computational bottlenecks associated with traffic calibration and expand beyond simple volume adjustments, several contemporary traffic models rely on metaheuristic approaches, such as GAs.

Prior relevant research rigorously validates the use of GAs for calibrating complex urban networks, including the integration of GAs directly with the SUMO environment \cite{khaleghian2023calibrating, pourabdollah2017calibration}. While this establishes a clear precedent for our specific simulation framework, the broader field of GA-based traffic optimization—across various simulation platforms—focuses on fine-tuning microscopic driver behavior to alleviate congestion. A substantial cluster of this research, including the two aforementioned studies involving SUMO, applies GA optimization to the calibration of specific parameters regarding specific car-following and lane-changing models. The methods minimize the error between simulated vehicle movements and real-world data by altering localized variables such as driver reaction times, minimum gap distances, acceleration profiles, among others \cite{yu2017calibration, kim2005calibration, esfahani2019new, dadashzadeh2019automatic, afshari2025calibrating, ma2007calibration, park2005development, paz2014development, ma2002genetic}. Beyond individual driver behavior, other applications within this domain expand GA optimization to broader network management parameters. For instance, GAs have been successfully deployed to optimize traffic signal schedules \cite{rahaman4multi} and to calibrate dynamic routing and path-choice parameters \cite{ma2002genetic}. Moving even further toward macroscopic demand calibration, some methodologies utilize GAs to directly tune aggregate vehicle input flows. This approach often relies on minimizing the discrepancy between simulated network states and real-world Floating Car Data (FCD) to iteratively adjust macroscopic volumes \cite{tettamanti2015iterative}.

While these studies successfully calibrate driver behavior, routing logic, and aggregate volumes, fundamental traffic theory emphasizes that accurately calibrating the underlying Origin-Destination (O-D) demand flows is equally critical to modeling network dynamics \cite{balakrishna2007calibration}. Yet, even when incorporating demand calibration, the existing literature predominantly couples it with the adjustment of driver behavior. Therefore, our methodology addresses a critical gap: we propose a framework that optimizes the spatial distribution of work-related demand rather than microscopic driver-behavior parameters. Rather than altering driver behavior, our proposed methodology focuses on the spatial distribution of job locations. By changing the distribution of jobs across road segments, the flow of traffic is inherently modified as the movement from home to work serves as a primary driver of peak-period mobility. Our GA is then able to optimize job locations, allowing ActivityGen and SUMO to produce traffic resembling true traffic patterns.

\section{Simulation Environment and Data Sources}
The methodology presented in this work relies on a SUMO-based simulation environment combined with real-world household and traffic-flow datasets. This section describes the study area, simulation framework, and data sources used throughout the calibration process.
\subsection{Study Area}
The study area used in this work consists of Greensboro, North Carolina, and the surrounding region. This area was selected due to the availability of a high-resolution SUMO-based traffic simulation environment developed through the GreenEVT framework \cite{nilsson2024greenevt}. The simulation includes a broad range of urban, suburban, and rural road environments and provides a useful environment for evaluating the ability of a GA to produce realistic traffic flows.

\subsection{SUMO}
The main simulation tool for this study was the Eclipse Simulation of Urban MObility tool, more commonly called SUMO, used in conjunction with a simulation tool called ActivityGen. Together, these tools handle the city road network, traffic routing, and the collection of traffic-flow information. ActivityGen is especially important because it allows home and job locations to be assigned along streets and dynamically routes work trips between them based on adjustable work-schedule settings. Traffic is simulated in accordance with the street conditions. For example, street signs and traffic control devices are obeyed.

Our work-schedule settings simulate a single workday, with worker start and end times left at the default ActivityGen values. We populated this city with 1000 people distributed across 500 households. Although this population is much smaller than the actual population of Greensboro, the evaluation compares relative traffic-flow patterns rather than absolute vehicle counts. For this reason, the reduced population size was not expected to substantially affect the correlation-based evaluation, while also making training less computationally expensive.

The job and housing distributions are encoded as normalized values distributed across the streets in SUMO, such that the relative values of given road segments determine what share of the total houses and jobs are allocated to any given road. The housing distribution is initialized at the start of the algorithm using known housing data and is then held fixed, while the job distribution is updated by the algorithm during training.

SUMO additionally allows entrances and exit points to be allocated for vehicles entering or leaving the simulation from outside the modeled road network. These are referred to as "gates" and  are distributed across the roads that intersect the boundary of the imported road network. We decided that both the number of vehicles entering and leaving through the gates, and the share of incoming cars that each gate receives were very likely to have meaningful impacts on the  quality of our traffic representations. As such, the GA updates three parameter groups during training: the job distribution, the total number of vehicles entering and leaving through gates, and the distribution of incoming and outgoing vehicles across those gates.

In this work, the housing distribution is treated as a fixed input, while the GA modifies the job distribution, total gate traffic, and gate-traffic distribution.

\subsection{Data Acquisition}
Because this paper is motivated by infrastructure planning, we wanted the data required to apply our model to be as generally available as possible. We would like this model to be applicable to as many cities as possible, and having difficult to acquire data be required for our pipeline would interfere with this goal. We tailored the inputs to our model around those we found to be generally easy to acquire for any given city.

Our system relies on an imported network of streets in our study area, a real-world home distribution to initialize our cities with, and some sample of streets with real-world traffic-flow data. 

The street network is imported from OpenStreetMap, which provides road-network data for many cities globally. These networks are imported directly into SUMO and include mapped traffic control devices, and are easy to import for most cities.

The real-world home distribution is derived from data published by the US Census Bureau, using the same traffic analysis zones (TAZs) that have been examined in previous work \cite{nilsson2024greenevt} \cite{census2010taz}. We assign homes to roads in SUMO by associating each road with its nearest TAZ and distributing the known number of households for that TAZ across the roads uniformly. 

The sample of streets with real-world traffic-flow data for our work comes from the North Carolina Department of Transportation published in 2024. Their data includes measurements of average annual daily traffic for a number of road segments within the city of Greensboro and the surrounding areas. This data only includes a small sample of the total number of roads or road segments that exist within the area examined by this paper. 

\section{Proposed Calibration Framework}
\subsection{Proposed Genetic Algorithm}
The general use of a genetic algorithm is to optimize a set of inputs such that a fitness value is maximized. The genetic algorithm used in this work optimizes towards a fitness value derived from correlation between simulated and real traffic on known roads. It does this by modifying three parameter groups: the distribution of work positions, the total number of vehicles entering and leaving through network gates, and the distribution of those vehicles across individual gates. These three feature sets are referred to as the "genome" of the algorithm. Our algorithm functions through stages of fitness evaluation, mutation, and crossover across a number of different genomes.

Each candidate solution is represented as a genome. In this paper, we refer to the 16 candidate genomes in each generation as children. Our algorithm functions by initializing 16 children that each have a randomly initialized genome, but are otherwise the same. For each generation of children, or set of 16 children undergoing mutation and crossover with each other, a simulation is run using the parameters inside of each child's genome. During this process the number of cars entering each road segment is stored and used to calculate fitness as a correlation between the number of cars that entered a segment and the average annual daily traffic value for that road from the North Carolina Department of Transportation. This fitness value is compared across the 16 children and the highest-fitness child is selected as the designated "parent" for the following generation. The parent is also not affected by the crossover or mutation stages.

The parent for a generation has its genome combined with the 15 other children during the crossover stage. The implementation of this in our work has to account for the three separate elements that each child's genome is composed of. Crossover for the job array is performed by selecting approximately 80\% of each non-parent child's streets, and averaging the work positions on these streets between the parent and the child. The distribution of incoming and outgoing traffic across the different gates has the same process applied to it, with the share of total incoming and outgoing traffic each gate receives being modified. 

The total number of incoming and outgoing vehicles underwent crossover differently. For every single child both the outgoing and incoming total number of cars underwent a weighted averaging between the parent and the child with slight random variations in the weighting. 

The mutation mechanism for the workplace and gate traffic distribution was similar to the crossover in that each specific street and job location had a probability of being mutated. The mutation operation consisted of a random multiplication by a float between 0 and 2, followed by an addition of a random value between -3 and 3 to avoid getting stuck at 0. This operation was performed on each selected road segment’s job-distribution value, and any resulting negative value was set to zero.

The mutation mechanism for the total incoming and outgoing traffic functioned by selecting a random proportion of the children and then multiplying the incoming and outgoing traffic values for both by a random value between 0.5 and 1.5.

For all three elements of the genome, there was a random proportion of either genes or children mutated. This random percentage was generated based on a sine function of the generation that oscillated between approximately 91\% and 8\%. This allows our model to more easily escape local optima while still allowing small changes within the solution space.


\section{RQ1: Full-Dataset Calibration Performance}
\subsection{Evaluation Method}
The first research question examines whether our proposed methodology improves the alignment between simulated traffic and real-world traffic. To evaluate this, the algorithm was trained for 200 generations on the full set of available road segments whose real-world traffic flow was measured. The correlation between simulated and observed traffic was then tracked across generations. We performed this training over 5 runs using multiple random seeds for initialization and genetic modification. 

\subsection{Results}
\begin{figure}[h]
    \centering
    \includegraphics[width=1.1\linewidth]{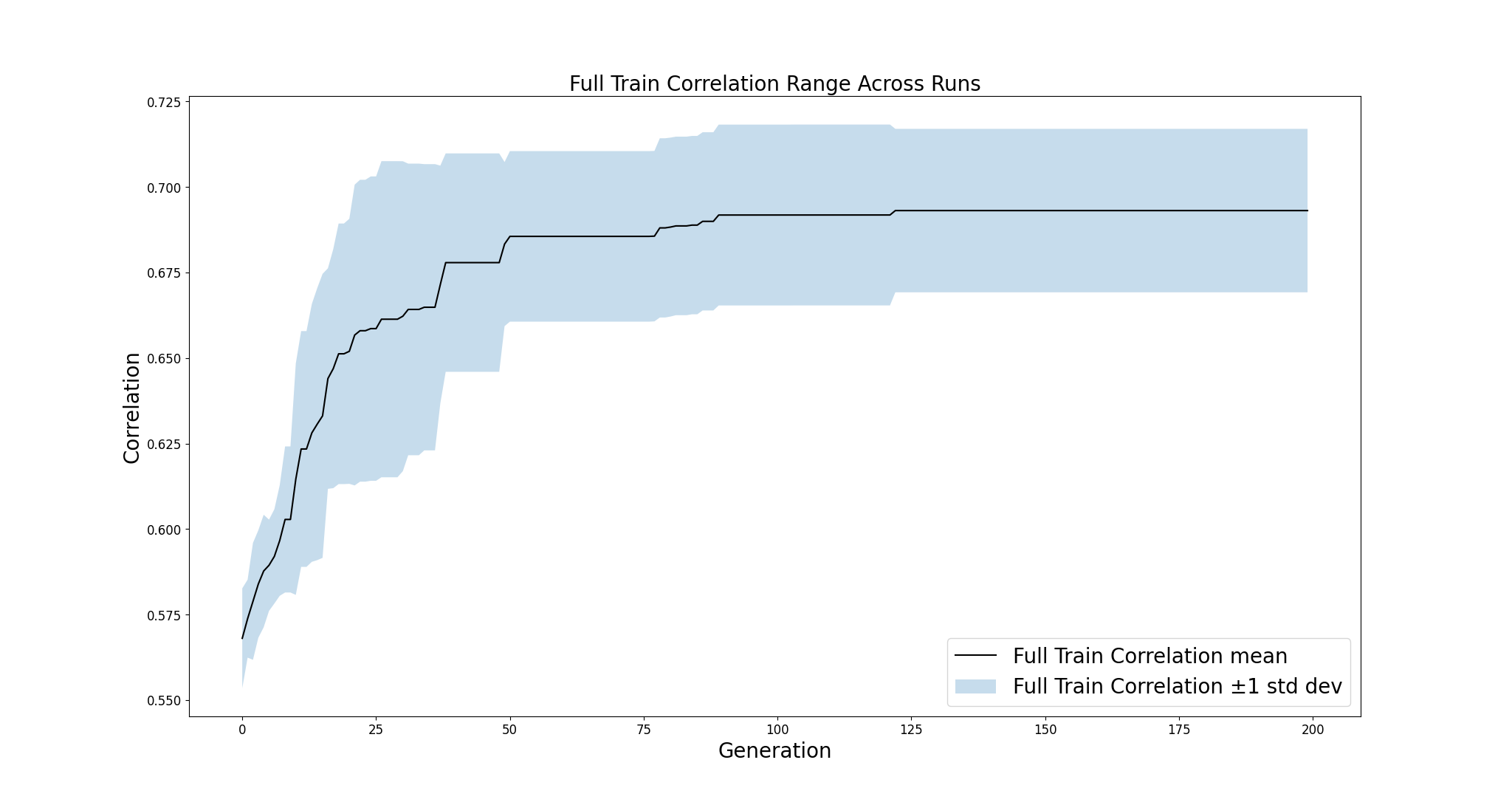}
    \caption{The mean and 1 standard deviation of correlations from the full training set.}
    \label{fig:0_percent_exclusion}
\end{figure}

The average correlations and 1 std deviation of the resulting correlations are seen in Fig. \ref{fig:0_percent_exclusion}. The correlation generally rises rapidly, then stabilizes to between approximately 0.64 and 0.72. This indicates that our algorithm succeeds in improving the alignment of simulated traffic with real traffic.

\subsection{Interpretation}
These results show a clear increase in alignment between simulated and real-world traffic when using the proposed training method. This supports the claim that our GA is successfully modifying job and gate-traffic parameters in a way that improves the simulated traffic-flow pattern.

\section{RQ2 AND RQ3: Generalization to Excluded Roads}

Generalization to unseen roads is critical to the utility of our approach, as the sparse data setting that motivates this work assumes that most roads will have no known traffic-flow rates at deployment time. RQ2 and RQ3 directly address generalization beyond the available data.
\begin{figure*}[t]
    \centering
    \includegraphics[width=1\textwidth]{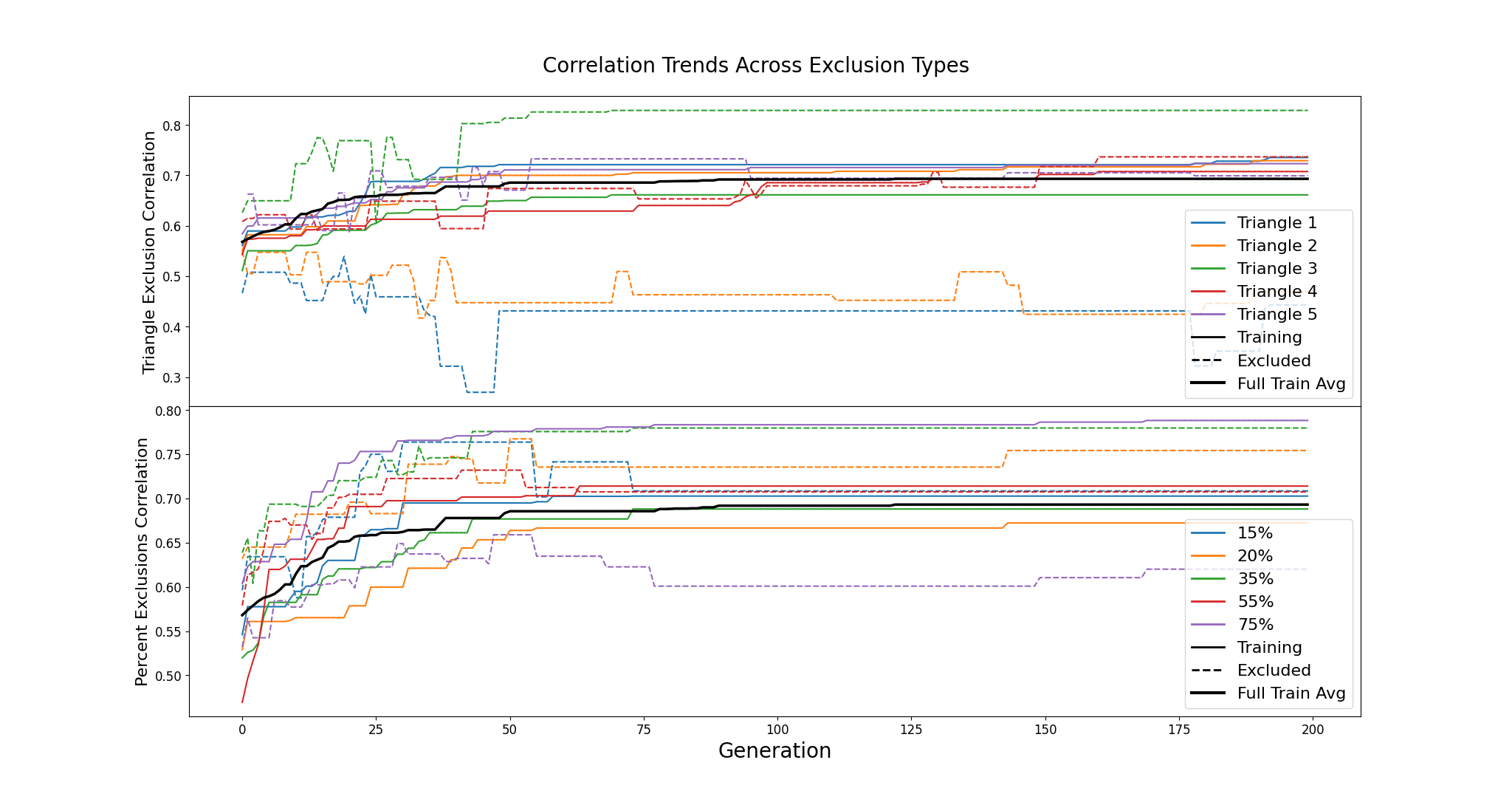}    
        \caption{The correlation over 200 generations for the percentage exclusion conditions and the geometric exclusion conditions}
    
    \label{fig:percent_exclusions}
\end{figure*}

\subsection{Evaluation Method}
Our second research question therefore examines whether road segments that are randomly excluded from training show improvement alongside the roads that are trained on. We evaluate this by selecting random subsets of 15\%, 20\%, 35\%, 55\%, and 75\% of the known road segments and excluding them from the training process. We train our model for 200 generations, and correlation with the real-world flow observations was measured separately for the roads that were trained on and the roads that were excluded.

\begin{figure}[h]
    \centering
    \includegraphics[width=\linewidth]{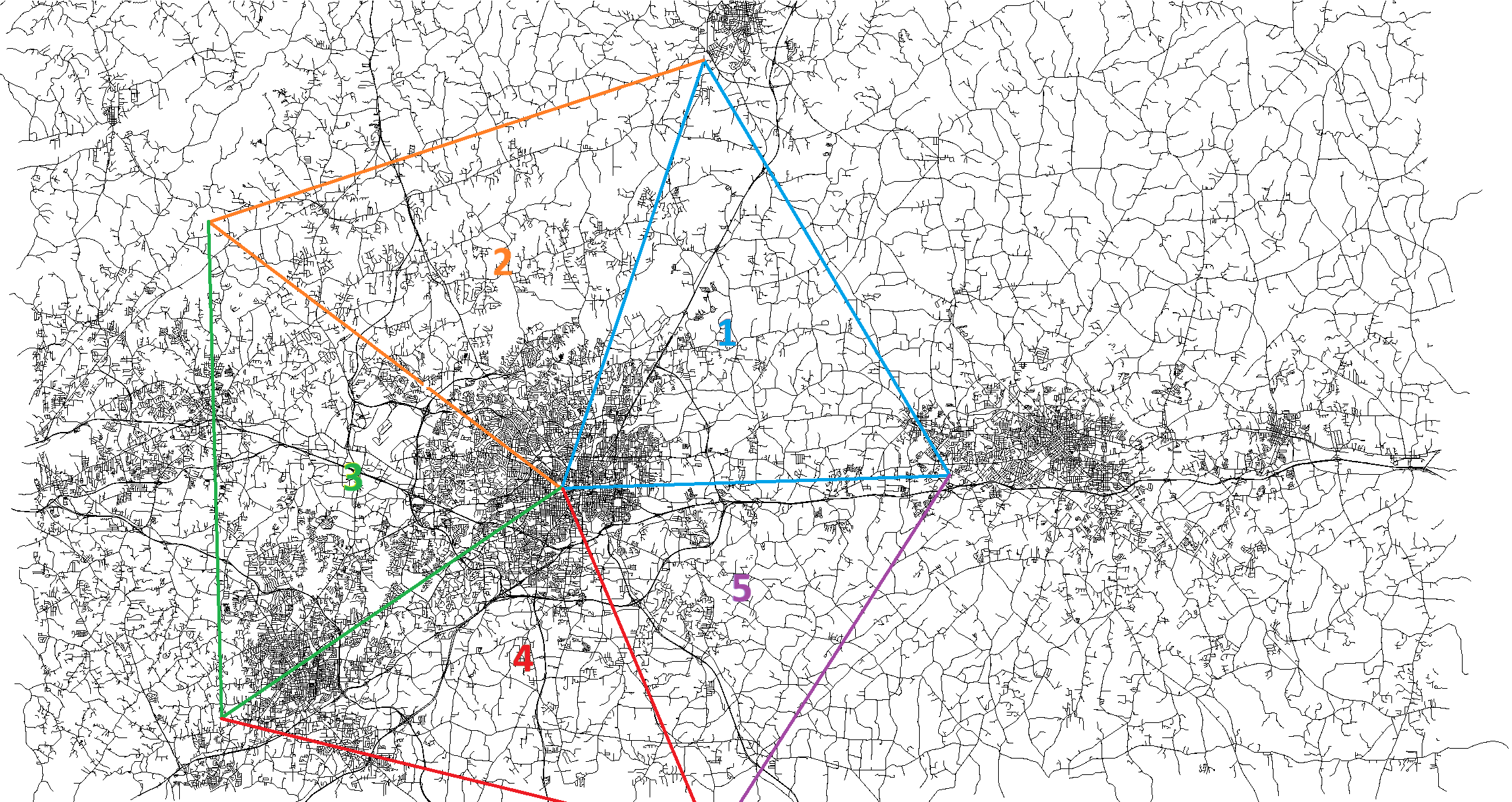}
    \caption{The triangular exclusion zones are roughly centered around the middle of the Greensboro city. A small portion of triangles 4 and 5 reside outside the simulation.}
    \label{fig:Triangles}
\end{figure}

The third research question examines whether geographically related regions that are excluded from training show improvement alongside the roads that are trained on. To evaluate this, large triangular exclusion zones were set within the simulation and removed from the training set. We again train over 200 generations, and measure correlations with the real-world data for both the trained roads and the excluded roads. The triangles are centered near downtown Greensboro City and can be seen in Fig. \ref{fig:Triangles}.

\subsection{Results}
The results of the training for both the exclusion triangle and percent based exclusion conditions can be seen in Fig. \ref{fig:percent_exclusions}. The results show that in many cases the excluded sets correlate more with real-world traffic than the training set. In other cases the excluded roads decrease in correlation as the training roads increase in correlation. 

Geographic exclusion represents a strictly harder generalization test than random exclusion, as it requires the model to infer traffic patterns in regions with no nearby training observations. Accordingly, triangular exclusion conditions produced less consistent results. One can observe that triangles 1 and 2 show very strong signs of overfitting, while triangles 3, 4 and 5 show training and excluded correlations both improving. 

Paradoxically, it can be seen that for the percentage exclusions, there appears to be a trend of increased correlations as the percentage of known roads that are excluded decreases.

\begin{figure*}[ht]
    \centering

    \begin{subfigure}[b]{0.325\textwidth}
        \centering
        \includegraphics[width=\textwidth]{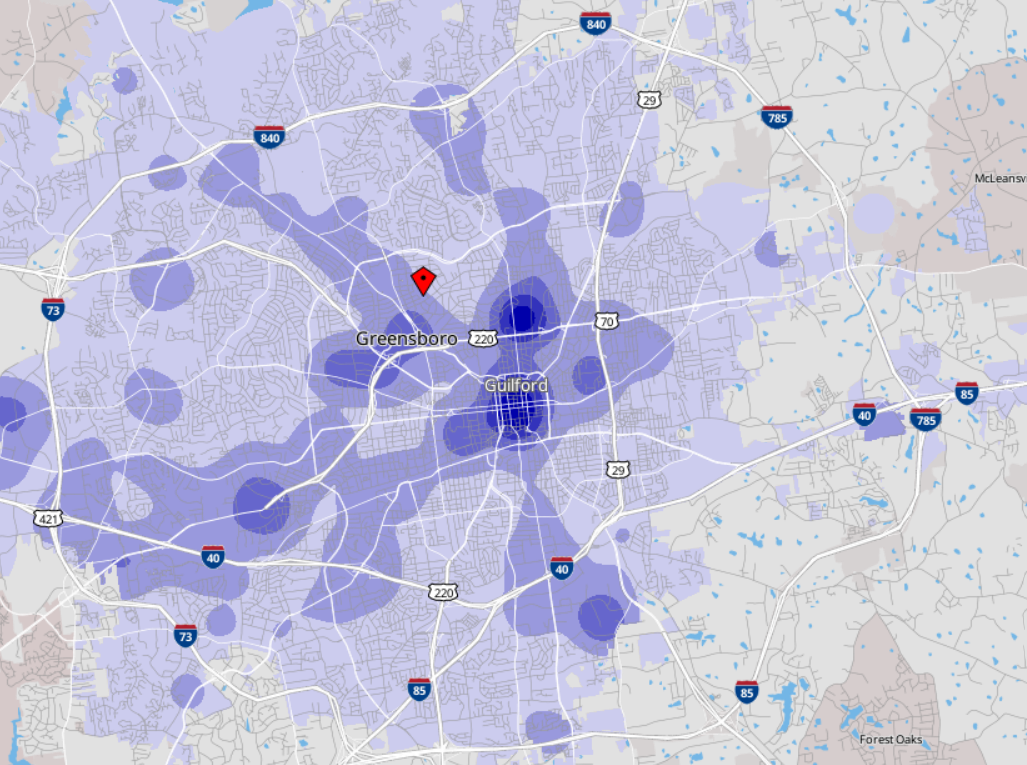}
        \caption{Real-World Job Distribution}
        \label{fig:first}
    \end{subfigure}
    \hfill
    \begin{subfigure}[b]{0.325\textwidth}
        \centering
        \includegraphics[width=\textwidth]{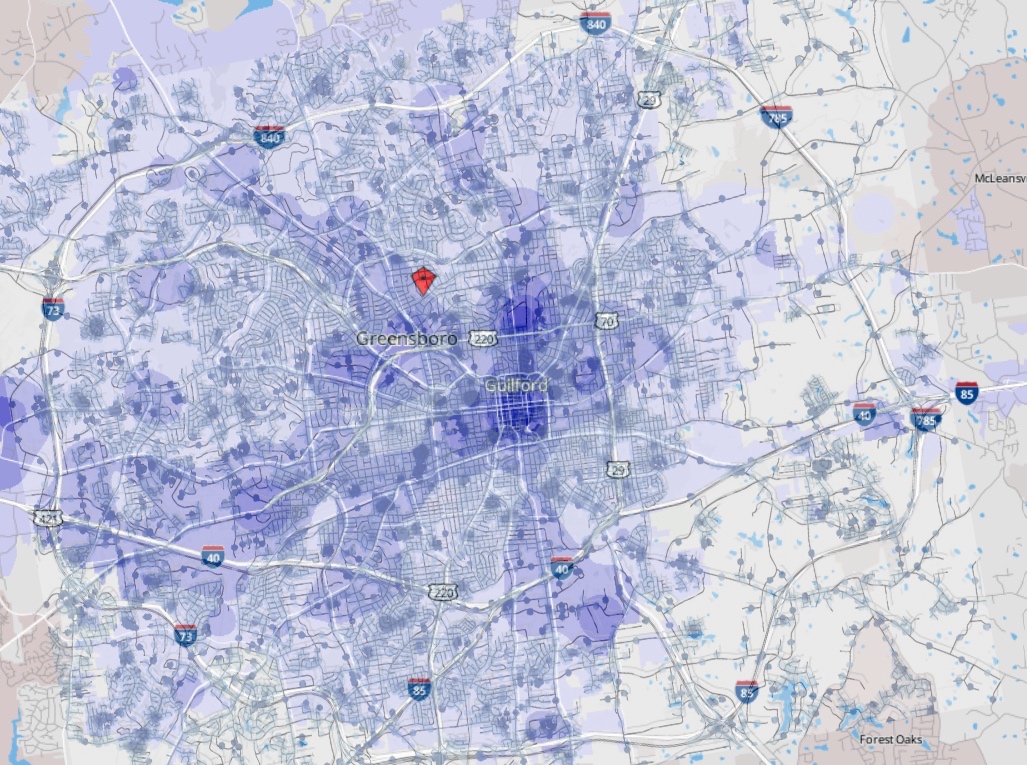}
        \caption{Overlaid Job Distribution}
        \label{fig:second}
    \end{subfigure}
    \hfill
    \begin{subfigure}[b]{0.325\textwidth}
        \centering
        \includegraphics[width=\textwidth]{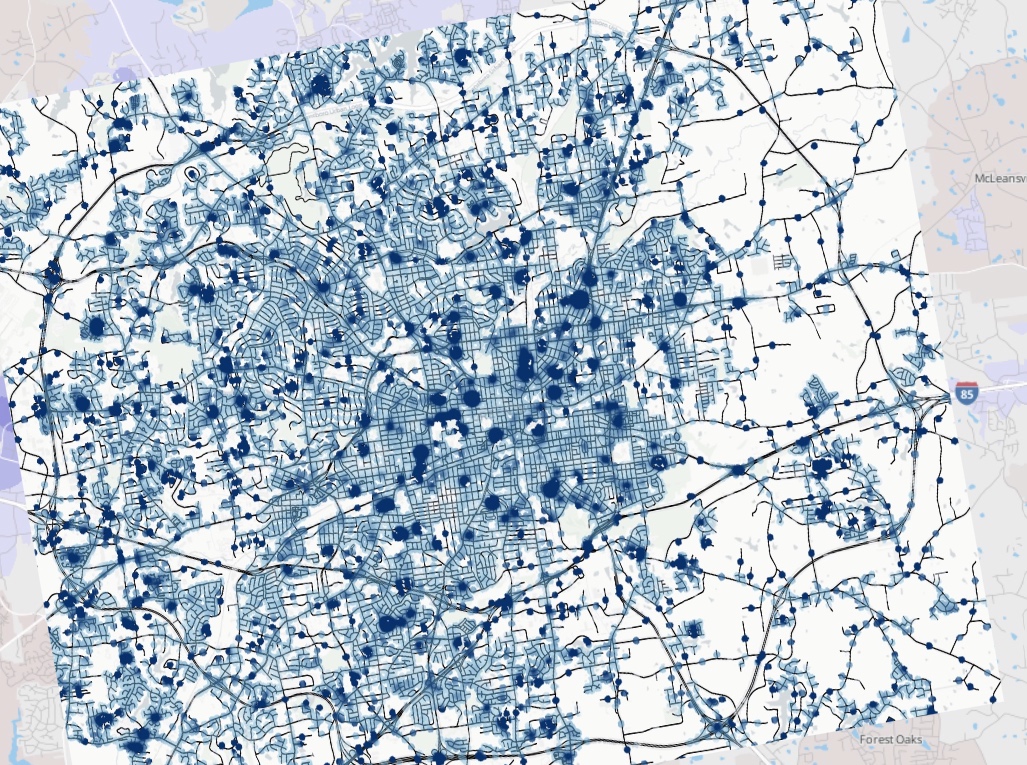}
        \caption{Simulated Job Distribution}
        \label{fig:third}
    \end{subfigure}

    \caption{Real-world job distributions as taken from the U.S. Census Bureau overlaid with simulated job distributions. Notable overlap is observed in the lowest and highest job density regions.}
    \label{fig:combined}
\end{figure*}

\subsection{Interpretation}
These results are somewhat counterintuitive, as we note that the roads we train to become more aligned with real-world traffic data do not consistently show higher correlation with real-world data than the set of roads that are excluded from training. We propose two main theories that explain this.

The simplest is that some roads are easier or harder to match to the real-world and simulated traffic, and the excluded roads have an overrepresentation of easy roads. For example, roads that are high traffic in the real world are likely high traffic because they are optimal paths for a number of possible routes. This could make them easier for them to produce stronger correlations with real-world data at higher levels for more possible genomes. If the training sets exclude many of these easier roads, the measured training correlation may decrease while the held-out correlation increases. It is plausible that this is occurring, especially since we do not know the ratio of "easy" to "difficult" roads in the dataset from the North Carolina Department of Transportation. 

Our other explanation examines the problem as a question of dimensionality of the solution space. Each element of the genome adds a dimension to the space in which solutions can exist, and excluding roads from the training set reduces the effective dimensionality of the optimization problem. If the solutions that exist in the larger and smaller solution spaces can be expected to have substantial agreement in their broader traffic-flow patterns, then reducing the total number of roads can simplify movement through the solution space. It could specifically be acting to filter out the impact of very localized traffic-flow patterns from the larger citywide traffic-flow structure. Because nearby roads and major traffic corridors are presumed to exhibit strong spatial correlation, these lower-dimensional solutions may still generalize well to the excluded roads, with a stronger resistance to idiosyncratic local traffic patterns.

For the percent exclusions, we note that the 75\% exclusion condition had a much higher training correlation than held-out correlation and the 55\% exclusion condition behaved similarly, although with a less pronounced effect. This may indicate that the generalization benefits associated with dimensionality reduction disappear once the training set becomes too small. If the dimensionality reduction explanation is correct this overfitting behavior could indicate a limit on the benefits provided by reducing the training set.

Our results also indicate that the generalization to unseen roads is much more consistent when missing observations are distributed randomly than when they are geographically clustered. This suggests that the calibration process depends on having some observed roads distributed throughout the study area. When an entire region is absent from the training set, the algorithm may fail to discover localized traffic patterns within the excluded region.

These results suggest that practitioners deploying this method should prioritize spatially distributed traffic observations over dense local measurements, as geographic coverage appears more important than observation density for achieving reliable generalization.

\section{RQ4: Comparison with Real-World Job Distributions}
\subsection{Evaluation Method}
The fourth research question examines how the job distributions generated by our GA compare with real-world job distributions. To evaluate this, heatmaps generated from the job distributions created by our model were compared visually against publicly available density maps derived from U.S. Census Bureau data \cite{census2010taz}.

We note that visual comparison is an inherently qualitative evaluation method. Quantitative validation of recovered job distributions across multiple cities remains an important direction for future work, and is a natural extension of the cross-city generalization agenda this work motivates.

\subsection{Results}

Job-density heatmaps for the Greensboro area are available through the U.S. Census Bureau. We generated a comparable heatmap from the optimized job distribution of the best-performing child in one full-training run. A comparison of these two can be seen in Fig. \ref{fig:combined}, with both shown on their own and then overlaid to compare clustering. The two maps show general overlap in the highest density pockets, and in shared areas of low numbers of jobs.



\subsection{Interpretation}

These results suggest that the job distributions generated by our model share meaningful similarities with real-world job distributions despite never directly training on that data. The observed overlap in both high and low density regions indicates that the traffic signal contains recoverable information about underlying urban employment structure, supporting the broader claim that sparse road observations can serve as a useful proxy for unavailable urban data.



\section{Discussion and Conclusion}

The central challenge motivating this work is that realistic urban traffic simulation requires data that is rarely available at the resolution or scale needed for most cities. Our results demonstrate that this challenge can be substantially addressed through genetic optimization against sparse road observations. By optimizing job distributions and gate-traffic parameters against a small sample of roads with known traffic-flow rates, our framework produces simulated traffic that correlates well with real-world measurements while requiring only the kind of data that is realistically available for most cities.

The full-dataset calibration results confirm that the genetic algorithm successfully improves alignment between simulated and observed traffic, establishing that the optimization approach is sound. More importantly, the generalization results demonstrate that this improvement is not limited to the roads used for training. In the random exclusion conditions, held-out road segments show improvement alongside training roads across a range of exclusion levels, suggesting that the optimization is recovering broad city-scale traffic structure rather than fitting to individual road segments. This is the critical result for practical deployment, as it suggests the method can produce realistic simulations even when traffic observations are available for only a small fraction of the road network.

The geographic exclusion results reveal an important limitation. When entire spatial regions are absent from the training data, generalization becomes less consistent, suggesting that the method depends on spatially distributed observations rather than dense local measurements. This has a direct practical implication: cities deploying this method should prioritize collecting traffic observations that are spread across the road network rather than concentrated in specific areas.

The qualitative agreement between recovered job distributions and census employment data, despite never directly training on that data, provides additional evidence that the traffic signal contains recoverable information about underlying urban structure. This is a secondary but encouraging finding that motivates future quantitative validation across multiple cities.

Taken together, these results establish that sparse road observations can serve as a useful signal for calibrating realistic urban traffic simulations, reducing the data requirements for simulation deployment to what is realistically available for most cities. This provides a foundation for scalable traffic simulation that we hope to build upon to support applications including electric vehicle demand modeling and charging infrastructure planning across diverse urban environments.

\section*{Acknowledgment}
This material is based upon work supported by the National Science Foundation under Grant No. 2347337. Any opinions, findings, and conclusions or recommendations expressed in this material are those of the author(s) and do not necessarily reflect the views of the National Science Foundation.

\bibliographystyle{IEEEtran}
\nocite{*}
\bibliography{Ref.bib}

\vspace{12pt}
\color{red}

\end{document}